\documentclass{article}

\usepackage{arxiv}

\usepackage{authblk}
\usepackage{array, booktabs, multirow, graphicx}
\usepackage{color}
\usepackage{soul}
\usepackage{wrapfig}
\usepackage{amsmath, amssymb}
\usepackage{hyperref}

\newcolumntype{M}[1]{>{\centering\arraybackslash}m{#1}}

\title{Curriculum-guided Change Detection Training: Toward Accurate Serac Fall Monitoring}
\date{May 29, 2026}

\author[1, 2]{Arthur Dérédel\thanks{\texttt{arthur.deredel@liris.cnrs.fr}}}
\author[1]{Carlos Crispim-Junior}
\author[2]{Pierre Lemaire}
\author[2]{Johan Berthet}
\author[1]{Laure Tougne Rodet}

\affil[1]{LIRIS, UMR5205, Université Lumière Lyon 2, CNRS, Ecole Centrale de Lyon, INSA Lyon, Université Claude Bernard Lyon 1\\ 69676, Bron, France\\}
\affil[2]{Styx4D, 19 rue lac Saint André, Le Bourget-du-Lac, 73370, France}

\def\eg{\emph{e.g}\oneDot}

\def\etal{\emph{et al}\oneDot}
\def\IoU{\mathrm{IoU}}

\begin{document}
\lhead{\scshape Dérédel et al.: Curriculum-guided Change Detection}
\chead{}
\maketitle

\begin{abstract}

Change Detection (CD) aims to identify semantic or structural changes from nearly registered multi-temporal images. While recent advances in training methodologies have largely focused on semi-supervised learning and consistency regularization, alternative training paradigms remain underexplored. In particular, most deep CD methods rely on uniform sampling during training, implicitly assuming that all training samples contribute equally to the optimization process. However, such naive sampling can introduce noisy gradients and hinder robust representation learning.
To address this limitation, we propose a curriculum learning framework tailored for change detection. Our approach investigates two complementary difficulty measures: the Solar Angular Gap (SAG), a physically grounded proxy for acquisition-condition variability, and the Structural Similarity Index Measure (SSIM), which evaluates appearance similarity between image pairs. Based on these criteria, the framework progressively introduces challenging samples during training, enabling models to learn robust representations in a coarse-to-fine manner.
We evaluate our method on the challenging SeracFallDet benchmark, where results demonstrate consistent improvements of the proposed approach over standard uniform-sampling strategies for both pixel-based and object-based approaches. These results highlight the potential of curriculum learning to improve robustness in deep change detection. Importantly, our training framework is orthogonal to existing CD architectures, making it readily applicable to a broad range of methods.

\end{abstract}

\section{Introduction}
Change Detection (CD) is a core task in computer vision, aiming to identify differences between a target and a reference image of the same scene, taken at a different times. This task is crucial for applications such as autonomous driving~\cite{VL_CMU_CD, ZeroSCD} where continuously updating the environmental map is necessary, urban planning~\cite{AnyChange} or even natural hazard monitoring~\cite{SeracFallDet} where early detection of subtle events can prevent catastrophic outcomes. 

However, a major challenge comes from the different acquisition conditions where undesired change referred as pseudo-change may occur. As a matter of fact, an outdoor pair of images acquired under different sun positions introduces confusion: shadow casting, apparent texture differences, unrelated to the actual change of interest. Such confounding factors signals that samples are not equally hard to learn from. Moreover, curriculum learning has shown that signal quality and the order the training samples are introduced matters~\cite{curriculum_learning_survey1, curriculum_learning_survey2}. By introducing easier samples, with smoother loss landscapes, the model first gets guidance toward more robust representation~\cite{curriculum_learning_survey1}. Yet, to our knowledge, curriculum learning has never been applied to the change detection task. Adapting to this task is non-trivial as curriculum learning needs two key components: a difficulty measure and a pace to introduce harder samples~\cite{curriculum_learning_survey1, curriculum_learning_survey2}. While the latter can be derived from existing works~\cite{bengio_2009}, the former must account for the unique challenges of change detection. 

To address this, we investigate the Solar Angular Gap (SAG, the difference in sun position between image acquisitions) and the Structural Similarity Index Measure (SSIM) as proxies for sample difficulty. We show that larger angular gaps correlate with increased difficulty due to illumination changes and shadow casting, thus providing an adequate difficulty proxy, while the latter is commonly used to compare images.
Our contributions are threefold:
\begin{itemize}
\item We propose the first curriculum learning framework for change detection, motivated by the absence of reliable difficulty metrics in this domain.
\item We propose a physically-grounded, annotation-free difficulty proxy based on solar angular gap, which enables curriculum learning in both labeled and unlabeled settings.
\item We provide extensive empirical validation in the context of serac fall detection showing consistent improvements across architectures spanning segmentation-based and object-based change detection.
\end{itemize}

\section{Related Work}
In this section, we first review existing CD research directions, such as architectural improvements and training methodologies. Then, we introduce the curriculum learning concept, its key components and its successful applications across diverse tasks.
\subsection{Change Detection}
Change detection is an active research field where the objective is to identify pixels corresponding to semantic or structural changes in co-registered or nearly registered image pairs. To do that, most research has focused on architectural improvement to enhance models performances. Early approaches leveraged fully convolutional neural networks~\cite{FCNCD, C3PO}, while recent work has shifted toward transformer-based architectures~\cite{ChangeFormer} or hybrid methods~\cite{BiT, DR_TANet}. Those methods, like most nowadays, follow a dominant standard architecture. It typically comprises: a siamese network to extract features from the input image pair, a fusion module (\eg, multi-scale absolute feature difference or concatenation) to emphasize discrepancies, and a decoder to classify each pixel either as changed or unchanged.

Recently, research has introduced additional challenges to the CD task, such as handling large view variations~\cite{CYWS, CYWSv2}. To address these, some works have extended change detection architectures from pixel-level to object-level, offering a trade-off between precise boundary definition and object localization accuracy.

Despite significant architectural progress, training methodologies (particularly for fully supervised settings) have received comparatively limited attention. The most notable advances in this area stem from semi-supervised learning, with the consistency regularization framework~\cite{SemiCD, UniMatch, SemiCD_VL, UniMatchV2}. These approaches leverage unlabeled data by enforcing consistent outputs under strongly augmented inputs. However, these methods uniformly sample training data, overlooking the potential of prioritizing more reliable samples. Easier samples, which likely provide steadier gradients signals, are not exploited, potentially limiting model representation accuracy. 

\subsection{Curriculum Learning}
Curriculum learning, introduced by Bengio \etal~\cite{bengio_2009}, draws inspiration from human learning behaviors by prioritizing easier samples early in training. The core idea is that samples does not contribute equally to optimization: easier samples can guide the model toward more robust representations. Harder samples are then introduced gradually, according to a predefined difficulty metric~\cite{bengio_2009, curriculum_learning_survey1, curriculum_learning_survey2}. This paradigm has achieved notable success across multiple domains, including: semantic segmentation, where curriculum pseudo-labeling helps mitigate class imbalance~\cite{FlexMatch, FreeMatch}, self-supervised learning, where it stabilizes early training stages~\cite{CurriculumMIM}, and image generation, where it improves the quality of generated images~\cite{CurriculumDiffusion, CurriculumGAN}.

The difficulty of samples is often determined using task-specific heuristics, such as the number of foreground pixels~\cite{NumberObject} in semantic segmentation or the the amount of noise (time step)~\cite{CurriculumDiffusion} in diffusion models. Alternatively, some studies leverage model predictions to estimate sample difficulty~\cite{CurriculumMIM, CurriculumDomainAdaptation}. For instance, embeddings from pretrained models have been used to clusters prototypes~\cite{CurriculumMIM} with difficulty defined by the normalized distance to the nearest prototype. Other approaches employ model ensembling to distinguish reliable from unreliable samples~\cite{CurriculumDomainAdaptation}. Additionally, the training loss of samples can be used to iteratively update their difficulty scores~\cite{AdaptiveCurriculum}. 

Another key component in curriculum learning is the pace to introduce more challenging samples. Some approaches rely on predetermined strategies, such as milestone-based (\eg, every 10 epochs), linear, or quadratic scheduling to gradually incorporate harder samples in the training set~\cite{curriculum_learning_survey1}. In contrast, adaptive strategies emerged, delaying the introduction of difficult samples until convergence on an easier training subset~\cite{CurriculumDiffusion}.

Despite its success in other domains, curriculum learning strategies remain underexplored in change detection. This gap may stem from the challenge of defining sample difficulty in a task where changes are often subtle, noisy and context-dependent. Our work addresses this limitation by introducing task-specific difficulty measures for change detection. We aim to exploit the guidance of easier samples to improve model robustness. 

\section{Methodology}
In this study, we address the problem of volumetric change detection from a pair of co-registered optical images acquired by a fixed monocular camera. A change is defined as a spatially connected region whose underlying scene geometry underwent a permanent, instantaneous structural modification between the two acquisitions. Specifically, appearance variations that do not correspond to a physical modification of scene structure (e.g., cast shadows, temporary occlusions, etc.) are treated as pseudo-changes and are explicitly excluded. Such pseudo-changes are not trivial to handle and often reflects the difficulty of a sample. Indeed, pairs in which pseudo-changes are absent or negligible are inherently easier to learn from, as the supervision signal is unambiguous. Conversely, pairs exhibiting strong illumination shifts, partial occlusions, or heavy shadow casting require the model to disentangle structural from non-structural variation, a substantially harder discrimination.

As discussed before, the choice of difficulty proxy in curriculum learning is often task-specific~\cite{CurriculumDiffusion, NumberObject} or designed for single-image tasks~\cite{CurriculumMIM, HowHard}. While pre-trained difficulty scorers~\cite{HowHard} (based on human response time) can be beneficial in domains lacking clear difficulty measures, they fail to generalize across modalities. In our case, the shift from single images to image pairs further invalidates such approaches, as the difficulty of a pair cannot be directly derived from individual image scores. To address this, we propose to define a task-specific difficulty measure for change detection. In the following, we introduce and investigate two proxies: the widely used Structural Similarity Index Measure~\cite{SSIM} (SSIM), which quantifies structural dissimilarity between image pairs, and the solar angular gap (SAG). 

\subsection{Difficulty proxies}

The Structural Similarity Index Measure (SSIM), introduced by Wang \etal~\cite{SSIM}, is a widely used metric for comparing images and evaluating reconstructed images quality~\cite{LatentIntrinsics, SAIL}. Originally designed to mimic human visual perception by capturing structural information, SSIM quantifies similarity based on luminance, contrast, and structure.

As this metric compare image contents, it is sensitive to any source of apparent visual difference (including weather-induced variations), which increase perceived dissimilarity. More critically for curriculum design, SSIM also responds to actual semantic or structural changes in the scene. A pair containing great amount of changes will receive a high difficulty score, despite being precisely an pertinent sample the curriculum should prioritize. This confusion is an inherent limitation of content-aware proxies. Nevertheless, change detection datasets often exhibit heavy class imbalance~\cite{SeracFallDet, VL_CMU_CD}, where the vast majority of pairs contains little or no relevant changes. In such cases, SSIM high scores will reflect differences in acquisition condition rather than scene change. 

On the other hand, in outdoor image acquisitions, the relationship between sun position and observed pixel intensities is can be modeled by the Bidirectional Reflectance Distribution Function~\cite{BRDF} (BRDF), which describes how surface reflectance varies with illumination direction and viewing angle. In change detection, the camera position is typically fixed or subject only to small residual misregistration between acquisitions. Thus, the viewing angle component of the BRDF remains approximately constant, and then, under similar atmospheric conditions, the sun direction becomes the dominant factor governing changes in perceived scene appearance. This effect is further amplified by cast shadows: as the solar elevation and azimuth shift between acquisitions, shadow boundaries displace, altering the apparent intensity and texture of the scene.

In addition to that, only few parameters are required to obtain the solar position. Given a coarse camera position $p$, the algorithms VSOP2000~\cite{VSOP2000} or Michalsky, Joseph J.~\cite{solar_position}, can compute the sun location for any timestamp $t_i$. 

\begin{wrapfigure}[8]{r}{0.3\textwidth}
    \centering
        \vspace*{-2.0\intextsep}
        \includegraphics[width=3cm]{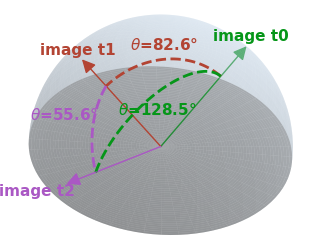}
        \caption{Solar angular gap between three timestamps.}\label{fig:solar_angular_gap}
\end{wrapfigure}

\begin{figure}[!b]
\centering
\rule{0pt}{1ex}\hspace{2.24mm}\includegraphics[width=12cm]{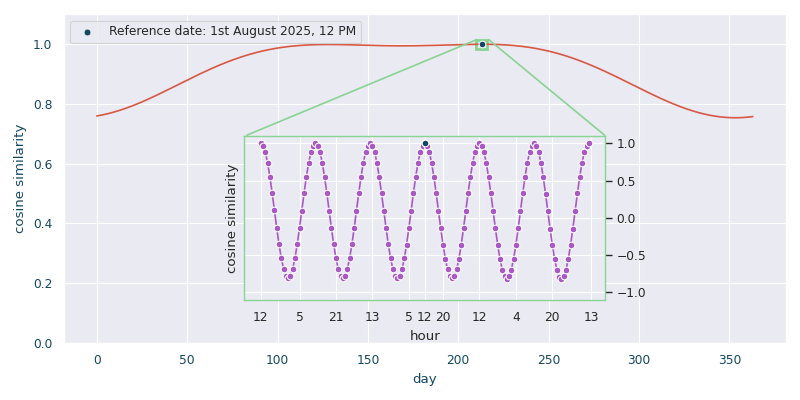}
\caption{Cosine similarity of the sun direction vector relative to a reference acquisition (1st August 2025, 12 PM, 45.88°N 6.85°E), computed over a full year at fixed hour (red) and over a $\pm$3-day window at varying hour (magenta). Intra-day direction variations substantially exceed inter-day variations.}\label{fig:cosine_similarity_evolution}
\end{figure}

Let $t_1$, $t_2$ be two timestamps of image acquisitions and $s_1, s_2 \in \mathbb{R}^3$ their respective solar directions vectors computed by the VSOP2000 algorithm~\cite{VSOP2000} at position $p$. The Solar Angular Gap (SAG) is defined as the cosine of the angle between the vectors $s_1$ and $s_2$: \begin{equation}SAG(s_1,s_2) = \frac{s_1 \cdot s_2}{||s_1||_2 \cdot ||s_2||_2} \end{equation}
The solar angular gap (SAG) between acquisitions, illustrated in~\hyperref[fig:solar_angular_gap]{Figure~\ref{fig:solar_angular_gap}}, thus directly governs some acquisition variations, independent to any semantic or structural change. As a physically grounded and image-content agnostic measure, SAG provides a robust proxy for acquisition dissimilarity. Intuitively, samples acquired at the same time of day are more likely to be similar, and is validated by the yearly evolution of solar direction which has less impact on the angular gap than hourly variations (see~\hyperref[fig:cosine_similarity_evolution]{Figure~\ref{fig:cosine_similarity_evolution}}).

While this measure offers a computationally efficient, label agnostic, and physically grounded difficulty proxy, it ignores weather-induced variations (\eg, cloud cover) or large shadows cast from surrounding landscape components, potentially limiting its effectiveness as a standalone difficulty measure.

\subsection{Curriculum pacing}
In standard deep learning training, batches are uniformly sampled from the entire dataset. In contrast, curriculum learning sorts this dataset by difficulty and then sample according to this ordering, with the goal of gradually introducing harder samples at a controlled pace. One common approach is the baby step pace~\cite{curriculum_learning_survey1}, where the dataset is split into $K$ clusters of increasing difficulty, $D = \{{D^{i}\}}_{i=1}^{K}$. During training, subsets $D^i$ are added at predefined milestones $m$ (\eg, epochs or iterations). Formally, the cumulative dataset $d(t;m)$ at iteration $t$ is given by:
\begin{equation}d(t; m) = \bigcup_{i=1}^{I(t; m)} D^i,\end{equation} where \begin{equation}I(t; m) = \max{\{j | m_j \leq t\}},\end{equation} is the index of the hardest subset included at iteration $t$. However, selecting milestones a priori is difficult, as the optimal pacing depends on the model's learning dynamics. 
To address this, we adopt the convergence-based pacing, following Kim~\etal~work~\cite{CurriculumDiffusion}: each subset $D^{i}$ is introduced only after the model has converged on $D^{i-1}$.  

Another explored alternative is to linearly increase the dataset size over time, starting from the easiest subset $D^0$ and progressively adding harder samples. In such case, the number of samples within the dataset $d(t;N)$ at iteration $t$ is given by: \begin{equation}|d(t; N)| = \min{(|D^0| + \frac{t}{N}(|D| - |D^0|), |D|)},\end{equation} where $ |D| $ is the total dataset size and $N$ is the total number of training iterations/epochs to introduce new samples. This approach provides a smoother transition between difficulty levels compared to milestone-based pacing.

\section{Experiments}
In this section, we present the experimental evaluation of curriculum learning for change detection. We first describe the SeracFallDet~\cite{SeracFallDet} dataset employed to evaluate our method. We then describe the experimental setup, including models, hyperparameters, augmentations and curriculum configurations. Then, we analyze relationship between the proposed difficulty metrics (SSIM and SAG) and sample difficulty and empirically evaluate their alignment with model performance, ensuring they effectively guide the curriculum. Finally, we report the results of our experiments, comparing the performance of curriculum learning over traditional uniform sampling. 

\subsection{Experimental setup}

\paragraph{Dataset.}{SeracFallDet~\cite{SeracFallDet} is a recently introduced change detection dataset focused on detecting serac falls (collapse of ice blocks in glaciers) from terrestrial time-lapse image pairs or sequences. This dataset contains considerable challenges: varying lighting conditions (natural illumination changes) and severe class imbalance where only $\sim$4\% of pixels are labeled as change in annotated samples. Following SeracFallDet~\cite{SeracFallDet}, we use two scenes with 552 image pairs as the training set, one scene with 140 image pairs for the validation set and the 5 remaining scenes with 425 image pairs are used as test set.}

\paragraph{Studied detectors.}{To validate our framework across both pixel-level and object-level change detection, we use two state-of-the-art architectures. The segmentation baseline is following the UniMatchV2~\cite{UniMatchV2} architecture. It consists in a siamese vision transformer (ViT-B)~\cite{ViT} with DINOv2~\cite{DINOv2} pretrained weights as encoder coupled with a DPT~\cite{DPT} decoder.
We adapt RT~DETR~\cite{RT_DETR} as the object-level baseline. The RT~DETR architecture consists in a siamese ConvNeXt-Base~\cite{ConvNeXt} with DINOv3~\cite{DINOv3} pretrained weights, followed by the RT~DETR~\cite{RT_DETR} neck and bounding box decoder. Both architectures fuse the extracted features using multi-scale feature absolute difference before the decoder.}

\paragraph{Hyperparameters tuning.}{Baseline hyperparameters are first tuned to maximize the performance on the entire training set (without curriculum). The curriculum variants then adopt the same configuration. This leads to a batch size of 8 with the AdamW~\cite{AdamW} optimizer (and a $1\mathrm{e}{-3}$ weight decay) for 300 epochs. Each epoch consists in 240 randomly selected training samples. Object-level change detectors were optimized with a frozen backbone and a learning rate of $5\mathrm{e}{-5}$ for the remaining weights. Pixel-level change detectors were trained under a $5\mathrm{e}{-6}$ and $2\mathrm{e}{-4}$ learning rate for the encoder and decoder respectively. The learning rates were decayed by a factor of 0.985 each epoch and each experiment used a single V100 GPU.}

\paragraph{Augmentations.}{To improve robustness, the training set is dynamically augmented using color jittering, blurring, random resizing (between 75\% to 100\% the original size). To address the class imbalance, each sample is cropped around a change mask (if present) or at a random location. Crop size is 518$\times$518 for pixel-level change detector and 640$\times$640 for object-level change detector. During validation and testing, the samples are tiled (without overlap) to match training crop resolutions.}

\paragraph{Curriculum configurations.}{We evaluate two pacing strategies with the proposed difficulty measures: baby step and linear paces. We used $K = 5$ different subsets for the baby step pace, trained until convergence on the validation set before integrating the following subset. As for the linear pace, harder samples are added every epoch for $N = 250$ epochs until reaching the whole dataset, starting with the whole first training subset $D^0$.}

\paragraph{Evaluation metrics.}{As both baselines are not designed for the same task, we use different metrics to evaluate pixel-level and object-level detectors. The pixel-level task is evaluated following the standard change detection pixel-wise metric Intersection over Union ($\IoU^c$)~\cite{SemiCD, UniMatchV2, SemiCD_VL} and F1\textsuperscript{c} score~\cite{SemiCD_VL}:
\begin{equation}\IoU^c = \frac{TP}{TP + FP + FN},\end{equation}
\begin{equation}F1^c = \frac{2TP}{2TP+FP+FN},\end{equation}
where TP, FP and FN indicate True Positives, False Positives and False Negatives respectively. Evaluating on the minority class better reflects model performances due to the class imbalance.

The object-level detection task is evaluated following an event-wise precision, recall and the corresponding F1\textsuperscript{c} as defined in~\cite{SeracFallDet}. Let $\mathcal{P}, \mathcal{G}$ be the sets of predicted bounding boxes and ground truths, respectively, and $\IoU^B$ denotes the bounding box intersection over union.
\begin{equation}
TP = |\{g \in \mathcal{G}, \exists p \in \mathcal{P} | \IoU^B(p, g) \geq \tau\}|
\end{equation}
\begin{equation}
FP = |\{p \in \mathcal{P}, \not\exists g \in \mathcal{G} | \IoU^B(p,g) \geq \tau\}|
\end{equation}
\begin{equation}
FN = |\{g \in \mathcal{G}, \not\exists p \in \mathcal{P} | \IoU^B(p,g) \geq \tau\}|
\end{equation}
where $\tau$ depicts the minimum $\IoU$ threshold (set as 25\%) to consider a prediction as true positive. We use these evaluation metrics to validate the effectiveness of our difficulty proxies (SSIM and SAG) in the following sections.}

\begin{figure}[!b] 
    \centering
    \includegraphics[width=10cm]{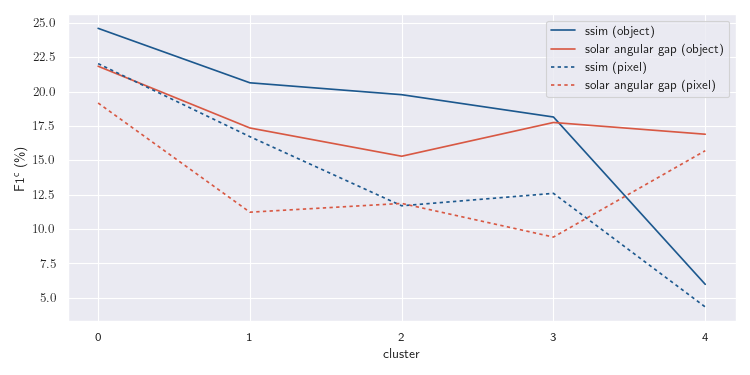}
    \caption{Performances (F1\textsuperscript{c}) across different cluster of difficulty on SeracFallDet~\cite{SeracFallDet} validation set.}\label{fig:score_across_difficulty}
\end{figure}

\subsection{Validation of difficulty proxy}
To assess the choice of the two proposed metrics (SSIM and SAG), we first needed to verify that they can be indeed viewed as a proxy of the difficulty of the task. For that purpose, we split the SeracFallDet~\cite{SeracFallDet} validation into five uniform difficulty clusters, binned by SSIM or SAG scores, and verify that the validation performance metrics are degraded by the level of difficulty. As illustrated in \hyperref[fig:score_across_difficulty]{Figure~\ref{fig:score_across_difficulty}}, the performance of the pixel-based baseline (trained on randomly ordered samples) performances generally declines as the model encounter more difficult samples, for both proxies. This confirms that both measures are strong candidates for assessing samples difficulty in this context. However, counterintuitively, performance recovers in the hardest SAG cluster. We hypothesize that it reflects annotator selection bias linked with a known limitation of the SAG proxy: under overcast conditions, clouds attenuate direct illumination, interfering into scene appearance. Despite this, the critical learning occurs during earlier training stages, which are more likely to approximate global minima~\cite{curriculum_learning_survey1, CurriculumTheory1, CurriculumTheory2}. Thus, SAG still fulfills its intended role in curriculum learning.

\begin{table}[!htbp]
    \centering
    \begin{tabular}{c c c c c}
    \toprule
    Pacing    & Difficulty            & Precision $\uparrow$ (\%) & Recall $\uparrow$ (\%) & F1\textsuperscript{c} $\uparrow$ (\%) \\\midrule
    Baseline  &                       & 16.8                      & \underline{27.8}       & 20.9                                  \\\midrule
    Linear    & \multirow{2}{*}{SAG}  & 21.6                      & 23.0                   & 22.3                                  \\\cmidrule{1-1} \cmidrule{3-5}
    Baby step &                       & 27.3                      & 26.5                   & \underline{26.9}                                  \\\midrule
    Linear    & \multirow{2}{*}{SSIM} & \textbf{30.2}             & 24.2                   & 26.8                      \\\cmidrule{1-1} \cmidrule{3-5}
    Baby step &                       & \underline{28.1}          & \textbf{29.5}          & \textbf{28.8}                         \\\bottomrule
    \end{tabular}
    \caption{\label{tab:object_level_validation}Object-level detector performances with different pacing and difficulty measures on SeracFallDet~\cite{SeracFallDet} validation set.}
\end{table}

\begin{table}[!htbp]
    \centering
    \begin{tabular}{c c c c}
    \toprule
    Pacing    & Difficulty            & IoU\textsuperscript{c} $\uparrow$ (\%) & F1\textsuperscript{c} $\uparrow$ (\%) \\\midrule
    Baseline  &                       & 14.0                                   & 21.7                                  \\\midrule
    Linear    & \multirow{2}{*}{SAG}  & 14.1                                   & 21.6                                 \\\cmidrule{1-1} \cmidrule{3-4}
    Baby step &                       & \textbf{17.4}                          & \textbf{24.6}                         \\\midrule
    Linear    & \multirow{2}{*}{SSIM} & \underline{16.1}                       & \underline{24.2}                      \\\cmidrule{1-1} \cmidrule{3-4}
    Baby step &                       & 15.3                                   & 23.3                                  \\\bottomrule
    \end{tabular}
    \caption{\label{tab:pixel_level_validation}Pixel-level detector performances with different pacing and difficulty measures on SeracFallDet~\cite{SeracFallDet} validation set.}
\end{table}

\subsection{Results}
\paragraph{Quantitative results.}{ 
We evaluate the impact of curriculum learning on both object-level and pixel-level change detectors, comparing them to their respective baselines (uniform sampling without curriculum learning), with different pacing strategies and the proposed difficulty measures.

\begin{table}[b!]
    \centering
    \begin{tabular}{c c c c c}
    \toprule
    Pacing    & Difficulty       & Precision $\uparrow$ (\%) & Recall $\uparrow$ (\%) & F1 $\uparrow$ (\%) \\\midrule
    Baseline  &                  & 20.6                      & \textbf{39.7}          & 27.1               \\\midrule
    Baby step & SSIM             & \textbf{32.8}             & 38.3                   & \textbf{35.4}      \\\bottomrule
    \end{tabular}
    \caption{\label{tab:object_level_test}Object-level detector results on the SeracFallDet~\cite{SeracFallDet} test set.}
\end{table}

All configurations improve performance over the baseline, except for the pixel-level detector~\hyperref[tab:pixel_level_validation]{(Table~\ref{tab:pixel_level_validation})} with SAG and linear pacing, which yields results similar to the baseline. The baby step pacing yields the most consistent and largest improvements on the validation set. As suggested by~\hyperref[tab:object_level_validation]{Table~\ref{tab:object_level_validation}}, most of the improvements in object-based change detectors imply the reduction of false positives, as the recall increases only marginally compared to the precision, effectively eliminating undesired pseudo-changes.

\begin{table}[!htbp]
    \centering
    \begin{tabular}{c c c c}
    \toprule
    Pacing    & Difficulty       & IoU\textsuperscript{c} $\uparrow$ (\%) & F1\textsuperscript{c} $\uparrow$ (\%) \\\midrule
    Baseline  &                  & 22.1                                   & 34.3                                  \\\midrule
    Baby step & SAG              & \textbf{25.1}                          & \textbf{37.6}                         \\\bottomrule
    \end{tabular}
    \caption{\label{tab:pixel_level_test}Pixel-level detector results on the SeracFallDet~\cite{SeracFallDet} test set.}
\end{table}

To assess the robustness of our training framework, we also verified that the best-performing configurations relatively to the baseline on the validation set also improved the results on the test set. These results show that not only curriculum learning does not degrade performance but can significantly improve the models robustness with a carefully chosen pace, achieving +8.3\% and +3.3\% F1 score improvement for object-level~\hyperref[tab:object_level_test]{(Table~\ref{tab:object_level_test})} and pixel-level~\hyperref[tab:pixel_level_test]{(Table~\ref{tab:pixel_level_test})} change detector respectively.}

\begin{figure}[t!]
    \centering
    \includegraphics[width=1.0\textwidth]{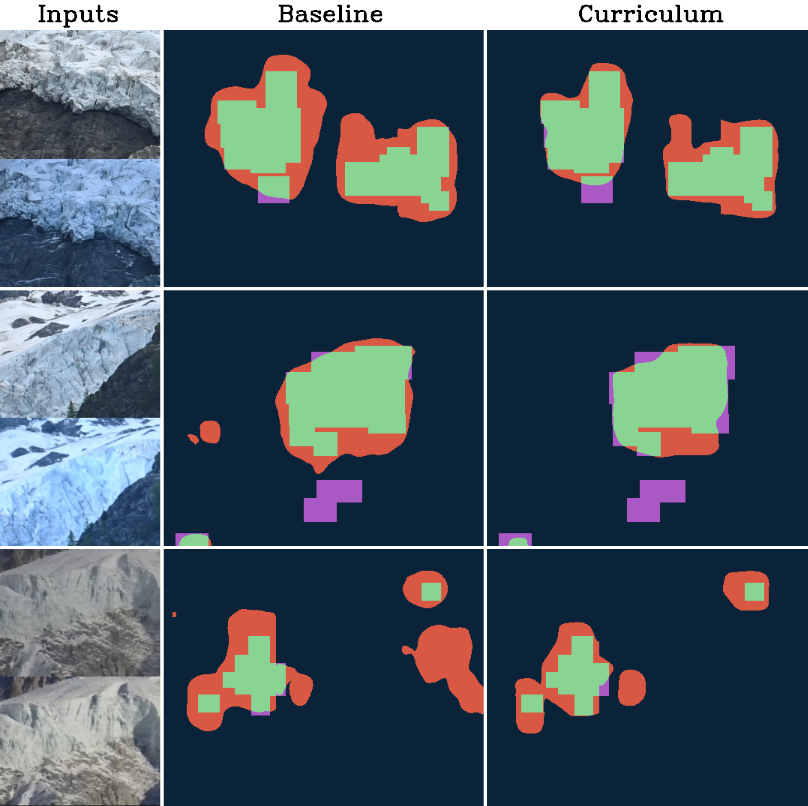}
    \caption{Pixel-level detector qualitative comparison between traditional training and curriculum learning. True positives, false positives and false negatives are indicated in green, red and magenta respectively. The left column represents the inputs pairs, the second and third column respectively represents the baseline and the baby step with SAG difficulty measure predictions.}\label{fig:qualitative_results_segmentation}
\end{figure}

\begin{figure}[t!]
    \centering
    \includegraphics[width=1.0\textwidth]{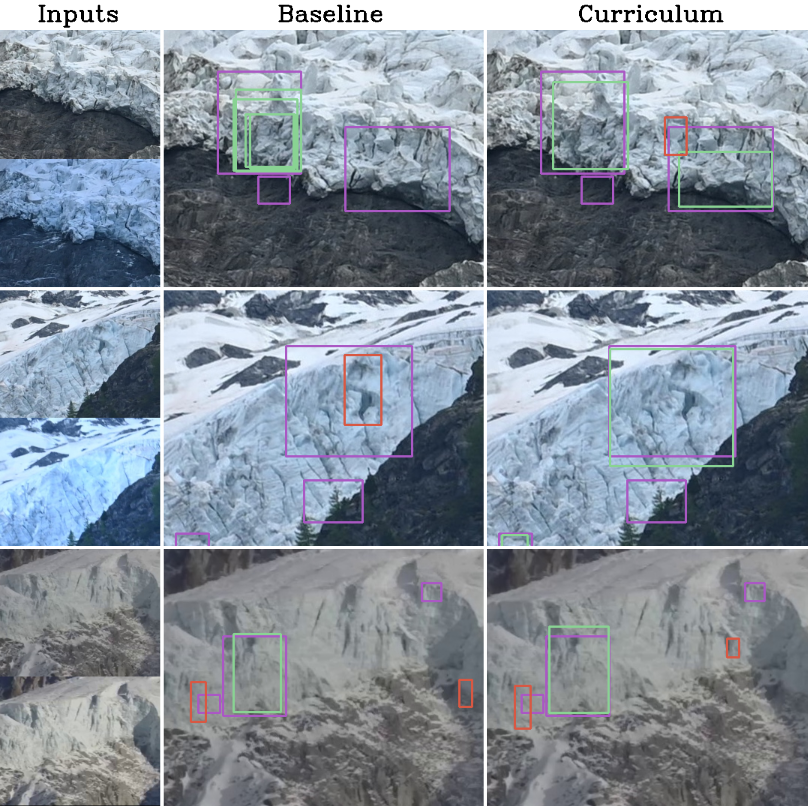}
    \caption{Object-level detector qualitative comparison between traditional training and curriculum learning. True positives, false positives and ground truths are indicated in green, red and magenta respectively. The left column represents the inputs pairs, the second and third column respectively represents the baseline and the baby step with SSIM difficulty measure predictions.}\label{fig:qualitative_results_object}
\end{figure}

\paragraph{Qualitative results.}{
While quantitative metrics provide a numerical assessment of model performance, qualitative results offer visual insights that are essential for interpreting why certain improvements occur and for identifying specific limitations.
\hyperref[fig:qualitative_results_segmentation]{Figure~\ref{fig:qualitative_results_segmentation}} illustrates the visual improvements achieved with curriculum learning on pixel-based approaches. Compared to traditional training, our method reduces false detections, resulting in less spread outputs. While some near boundaries may also be missed, this is likely due to noisy annotation (polygonal boxes instead of pixel-level labels), which exclude boundary pixels from ground truths. Thus, curriculum learning improves localization precision compared to traditional training, with sharper boundaries around the regions to detect.

These observations also hold for object-based approaches. As shown in~\hyperref[fig:qualitative_results_object]{Figure~\ref{fig:qualitative_results_object}}, the curriculum learning improved change event localization and reduced duplicate predictions for the same ground truth compared to the baseline (without curriculum).
}

\section{Conclusion and Future Work}

In this study, we demonstrated that curriculum learning consistently improves robustness across both pixel-level and object-level architectures, establishing it as an effective addition to change detection training pipeline. Central to curriculum learning framework, we propose two measures for the difficulty score: SSIM and SAG.\@ 
While the former involve comparing images content, the latter is content-agnostic (with minimal requirements), thus, it could be applicable in a broad range of outdoors acquisition settings.
Since SSIM and SAG displayed complementary weaknesses, SSIM being sensitive to actual changes while SAG is missing atmospheric acquisition condition, a path forward could combine both measures to mitigate individual weaknesses.
Beyond the specific proposed proxies, this work is orthogonal to most change detection methodology. A promising and most immediate direction is the extension to the semi-supervised settings, as SSIM and SAG are label-free. Finally, the performance gains linked to the SAG difficulty measure also indicate that lighting and volumetry play a key role in our problem. Thus, a promising direction of research could involve photometric stereo concepts, which exploit already the BRDF, to learn from unlabeled images.

\bibliographystyle{unsrt}
\bibliography{references}
\end{document}